\documentclass[conference]{IEEEtran}
\IEEEoverridecommandlockouts

\usepackage{cite}
\usepackage{amsmath,amssymb,amsfonts}
\usepackage{algorithmic}
\usepackage{graphicx}
\usepackage{textcomp}
\usepackage{subcaption}
\usepackage{xcolor}
\usepackage{hyperref}

\usepackage{color, colortbl}
\usepackage{xcolor}

\definecolor{LightBlue}{RGB}{212, 250, 252}

\usepackage{tikz}
\newcommand\copyrighttext{%
  \footnotesize \textcopyright 2024 IEEE.  Personal use of this material is permitted.  Permission from IEEE must be obtained for all other uses, in any current or future media, including reprinting/republishing this material for advertising or promotional purposes, creating new collective works, for resale or redistribution to servers or lists, or reuse of any copyrighted component of this work in other works.}
\newcommand\copyrightnotice{%
\begin{tikzpicture}[remember picture,overlay]
\node[anchor=south,yshift=10pt] at (current page.south) {\fbox{\parbox{\dimexpr\textwidth-\fboxsep-\fboxrule\relax}{\copyrighttext}}};
\end{tikzpicture}%
}

\def\BibTeX{{\rm B\kern-.05em{\sc i\kern-.025em b}\kern-.08em
    T\kern-.1667em\lower.7ex\hbox{E}\kern-.125emX}}
\begin{document}

\title{Towards Adversarial Robustness of Model-Level Mixture-of-Experts Architectures for Semantic Segmentation}

\author{\IEEEauthorblockN{Svetlana Pavlitska$^{1,2}$, Enrico Eisen$^{2}$, J. Marius Z\"ollner$^{1,2}$}
\IEEEauthorblockA{$^{1}$\textit{FZI Research  Center  for  Information  Technology}\\
$^{2}$\textit{Karlsruhe Institute of Technology (KIT)}\\
Karlsruhe, Germany \\
pavlitska@fzi.de}
}

\maketitle
\copyrightnotice
\thispagestyle{empty}
\pagestyle{empty}

\begin{abstract}
Vulnerability to adversarial attacks is a well-known deficiency of deep neural networks. Larger networks are generally more robust, and ensembling is one method to increase adversarial robustness: each model's weaknesses are compensated by the strengths of others. While an ensemble uses a deterministic rule to combine model outputs, a mixture of experts (MoE) includes an additional learnable gating component that predicts weights for the outputs of the expert models, thus determining their contributions to the final prediction. MoEs have been shown to outperform ensembles on specific tasks, yet their susceptibility to adversarial attacks has not been studied yet. In this work, we evaluate the adversarial vulnerability of MoEs for semantic segmentation of urban and highway traffic scenes. We show that MoEs are, in most cases, more robust to per-instance and universal white-box adversarial attacks and can better withstand transfer attacks. Our code is available at \url{https://github.com/KASTEL-MobilityLab/mixtures-of-experts/}.
\end{abstract}


\section{Introduction}

Deep neural networks possess several inherent insufficiencies~\cite{houben2022inspect}. In particular, they are vulnerable to deliberately generated adversarial noise~\cite{szegedy2013intriguing,goodfellow2014explaining}. One of the known defense strategies against adversarial attacks is to use a combination of neural networks, such as an ensemble. The idea is rooted in the reciprocal compensation of vulnerabilities: attacking an ensemble of multiple models, which produce different outputs for the same input, is more difficult than targeting a single model. Even if each individual model is vulnerable to attacks, the ensemble becomes robust because the strengths of others compensate for the weaknesses of one model. Several existing works have already demonstrated the efficacy of ensembling approaches as mitigation against adversarial attacks~\cite{abbasi2017robustness,kariyappa2019improving,pang2019improving}.

\begin{figure}[t]
\centering
\begin{subfigure}[t]{\columnwidth}
  \includegraphics[width=1.0\textwidth]{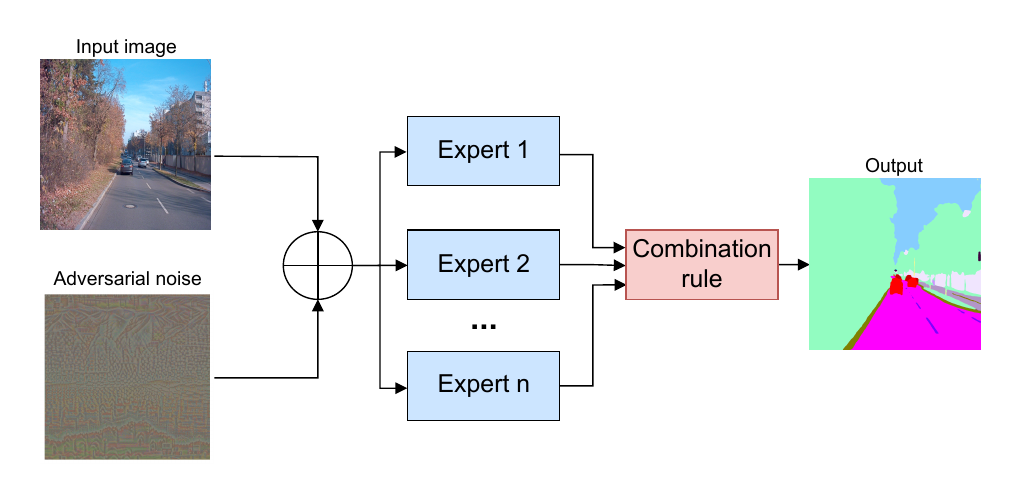}
  	\caption{Ensemble combines pre-trained expert models with a deterministic rule.}
\end{subfigure}
\begin{subfigure}[t]{\columnwidth}
  \includegraphics[width=1.0\textwidth]{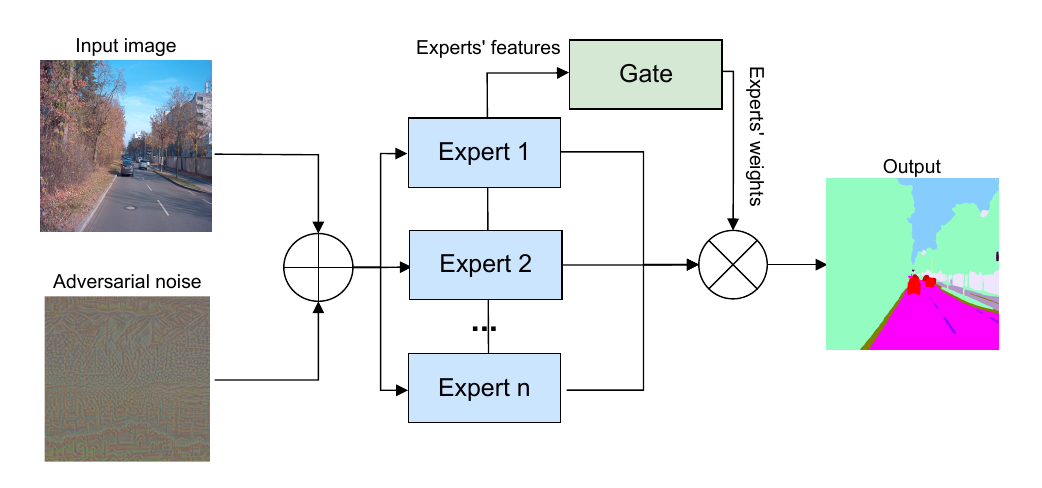}
  	\caption{Mixture of pre-trained experts uses a learnable gate component to predict expert contributions to the final prediction.}
\end{subfigure}
    \caption{Adversarial attacks on an ensemble vs. on a MoE.}
    \label{fig:concept}
\end{figure}

A mixture of experts (MoE)~\cite{jacobs1991adaptive} extends an ensemble with an additional gate component, which learns to predict the input-wise weighting of the expert sub-models. Differently from an ensemble, the combination of expert outputs may thus be different for each data input (see Figure~\ref{fig:concept}). A mixture of experts can be implemented at the model level, where individual experts are pre-trained models, or at the layer level, where experts are smaller network elements, such as layers\cite{shazeer2017outrageously}, residual blocks~\cite{pavlitska2023sparsely}, or channels in convolutional layers~\cite{wang2019deep}. While adversarial robustness of the sparse MoEs at the layer level has already been addressed recently~\cite{puigcerver2022on,zhang2023robust}, that of model-level MoEs has received no attention so far.

In this work, we aim to close this gap and analyze the adversarial robustness of a mixture of experts as a combination of pre-trained models. As an exemplary architecture, we consider an MoE for the semantic segmentation of traffic scenes from our previous works~\cite{pavlitskaya2020using,pavlitskaya2022evaluating}. This MoE consists of two experts specialized in \textit{urban} and \textit{highway} subdomains of input. We show that MoEs demonstrate a smaller drop in accuracy under per-instance and universal attacks than ensembles and are more robust to transfer of attacks among similar models. 

\newpage
\section{Related Work}
\label{sec:related}

\subsection{Mixtures of experts}

The MoE architecture was initially proposed by Jacobs et al.~\cite{jacobs1991adaptive}. An MoE consists of multiple specialized sub-models (experts) and a gating network that dynamically selects which experts to use for a given input. Each expert is trained to handle a specific subdomain of the input space, and the gating network assigns weights to the outputs of these experts based on the input data, effectively routing the input to the most appropriate experts. This differs from an ensemble approach, where the predictions of the multiple models are typically combined in a fixed manner, such as averaging or voting, without dynamic selection based on the input (see Figure~\ref{fig:concept}). Differently from this model-level approach, where experts are pre-trained models, an MoE can also be implemented at the layer level, where layers or larger neural network components serve as experts~\cite{shazeer2017outrageously}. This approach helps to drastically increase neural network capacity without a corresponding rise in inference speed. While the layer-level MoEs were extensively used for natural language processing tasks~\cite{fedus2022switch,rajbhandari2022deepspeed}, the model-level MoEs were primarily popular in the vision applications~\cite{ahmed2016network,pavlitskaya2020using,pavlitskaya2022evaluating}.

\subsection{Adversarial attacks on semantic segmentation models}

The vulnerability of semantic segmentation models to attacks with invisible adversarial noise has been considered in several works. Early work by Fischer et al.~\cite{fischer207adversarial} applied the least-likely method by Kurakin et al.~\cite{kurakin2017adversarial} to change predictions of a specific class to another class. In particular, the target class was set to be the nearest neighbor. 

The work by Arnab et al.~\cite{arnab2018on} was the first to apply adversarial attacks to the semantic segmentation task. With FGSM~\cite{goodfellow2014explaining} and its iterative variants, the authors performed per-instance attacks on images from the Pascal VOC~\cite{everingham2010pascal} and Cityscapes~\cite{cordts2016cityscapes} validation datasets. Models with residual connections (ResNet, E-Net, ICNet) demonstrated better adversarial robustness to attacks compared to VGG-based networks, although the latter possess a far larger capacity. Deeplabv2 with multiscale ASPP (atrous spatial pyramid pooling) was the most robust model during the evaluation. Also, multiscaling was shown to improve adversarial robustness.

Metzen et al.~\cite{metzen2017universal} addressed universal attacks on semantic segmentation models. The authors proposed two methods to perform targeted attacks. In the static target generation approach, a fixed target segmentation is defined. In contrast, dynamic target segmentation is designed to adapt to scene changes caused by ego motion. In this method, the segmentation stays unchanged except for certain classes, which are removed and replaced with target classes using a nearest-neighbor search.

Attack transferability for semantic segmentation models was explored by Gu et al.~\cite{gu2021adversarial}. They have evaluated PSPNet~\cite{zhao2017pyramid}, DeepLabv3~\cite{chen2017rethinking}, and FCN under FGSM and BIM~\cite{kurakin2017adversarial} attacks. They showed that the transferability between semantic segmentation models is limited compared to the image classification. They further showed that the transferability can be increased via a large number of attack iterations and random dynamic scaling of input and ground-truth data.

\subsection{Adversarial robustness via model combination}

An increase in adversarial robustness via model combination for the image classification task was first addressed by Abbasi and Gagné~\cite{abbasi2017robustness} for MNIST~\cite{metzen2017universal} and CIFAR-10~\cite{krizhevsky2009learning}. They have defined a subset of classes with high and low confusion using confusion matrices under FGSM attack. For each of these subsets of classes, they trained a specialist CNN on a subset of training data comprising corresponding classes. One generalist CNN trained on the complete data was also used in the resulting so-called \textit{specialist+1} ensemble. For the final prediction, the majority voting mechanism was used. The proposed \textit{specialist+1} ensemble was shown to better handle adversarial data generated with FGSM and DeepFool~\cite{moosavi2016deepfool}. 

Strauss et al.~\cite{strauss2017ensemble} evaluated the adversarial robustness of different ensemble types, including an ensemble of similar networks trained with random initial weights, an ensemble of various network architectures, bagging, and adding Gaussian noise to the training data to train models on slightly different datasets. Ensembles of ten classifiers were evaluated on MNIST and CIFAR-10 against FGSM and BIM attacks. Overall, all evaluated ensemble methods outperformed single classifiers on adversarial inputs. It was shown that random initialization of network weights leads to higher adversarial robustness than single classifiers. Also, two ways to generate an adversarial attack for an ensemble were evaluated: by using a gradient of one expert and the average of the gradients. For both datasets, FGSM attacks using the mean of the gradients were more successful, but for BIM, stronger attacks were achieved when using gradients of the one model out of the ensemble. Furthermore, bagging performed better on adversarial data but slightly worse on clean data.

An inherent characteristic of an ensemble is its diversity, i.e., the variability in model predictions. It has been shown that higher diversity not only leads to better performance~\cite{dietterich2000ensemble} but also results in higher adversarial robustness. This effect was first addressed by Pang et al.~\cite{pang2019improving}, later also in \cite{kariyappa2019improving,mehrtens2022improving}. Heidemann et al.~\cite{heidemann2021measuring} studied different diversity metrics and showed that especially the cosine similarity and double fault measure correlate highly with robustness metrics.

The mixture of robust experts~\cite{cheng2021mixture} was proposed to mitigate the attacks and adverse weather conditions by combining experts of three types:  clean experts, adversarially trained robust experts, and experts targeted at generated fog and snow. The experts, as well as the gate, followed the ResNet-18 architecture. To overcome the obfuscated gradients problem, the gate is trained together with the last FC layers of the experts. The proposed method outperformed other ensembling methods on the evaluated attacks.
Further approaches close to the MoE are the synergy-of-experts~\cite{cui2022synergy} and the immune MoE by Han et al. \cite{han2024enhancing}, both for the image classification task. Overall, the adversarial robustness of MoEs has remained less explored so far.

\begin{figure*}[t]
\centering
    \begin{subfigure}[t]{\textwidth}
    \centering
      \includegraphics[width=0.77\textwidth]{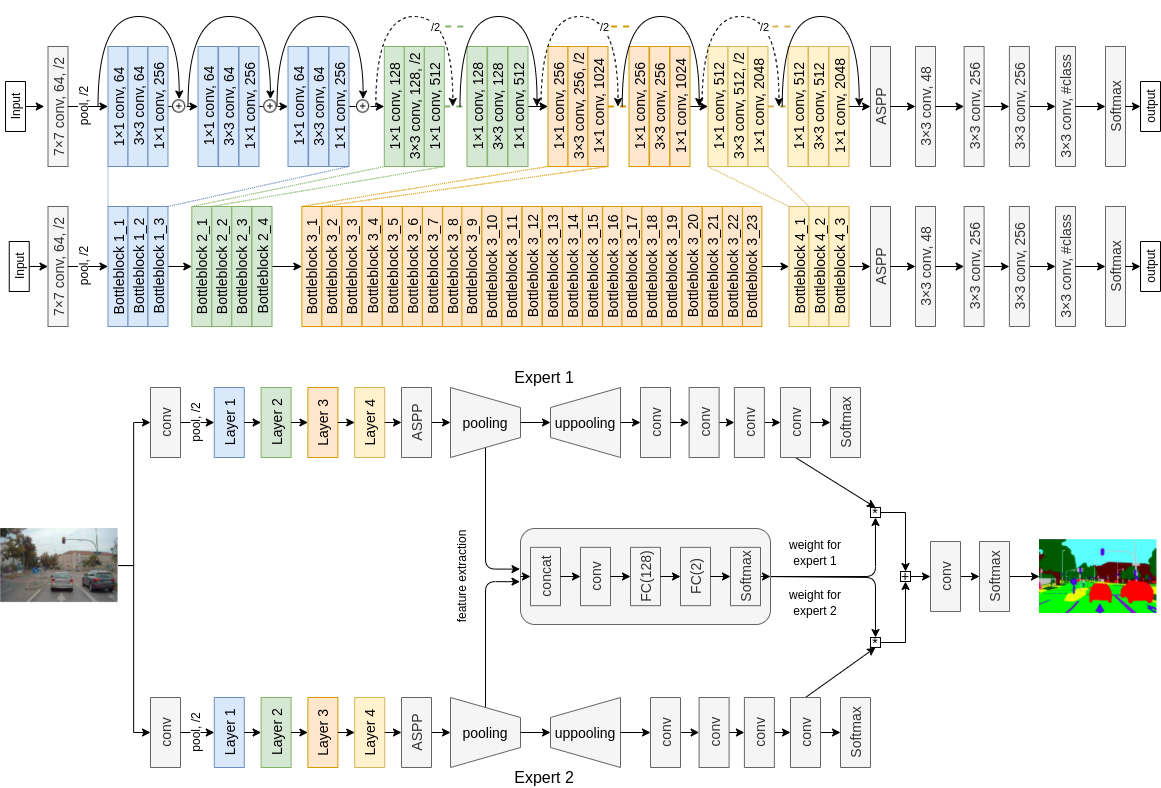}
      	\caption{\texttt{DeepLabv3$+$}-based MoE with ResNet-101 backbone.}
    \end{subfigure}
    
    \begin{subfigure}[t]{\textwidth}
    \centering
      \includegraphics[width=0.7\textwidth]{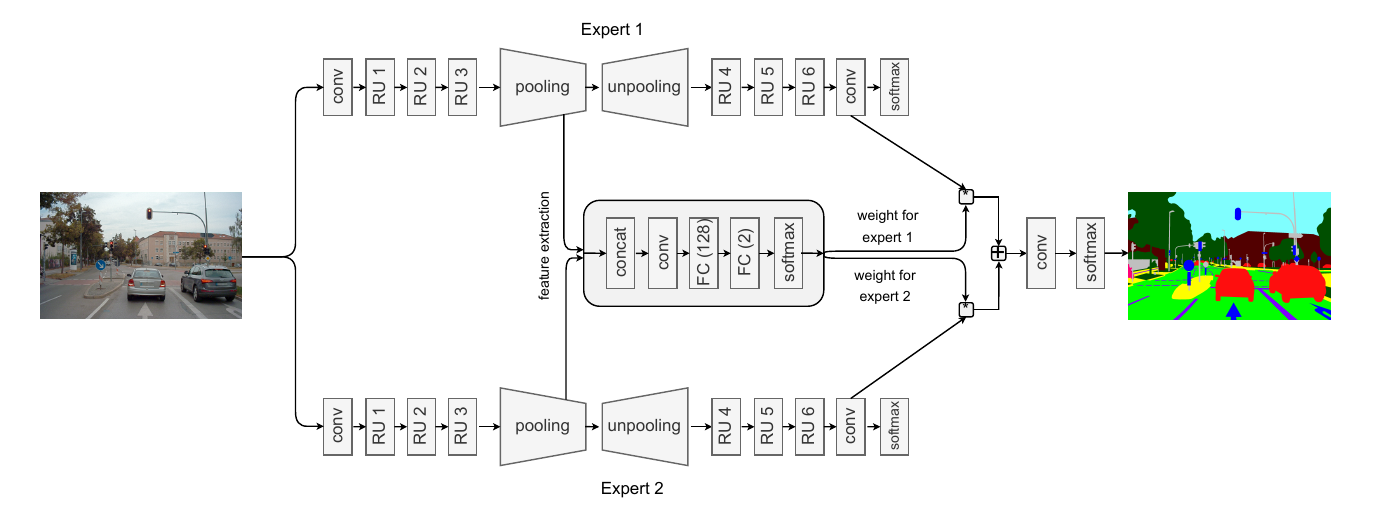}
      	\caption{\texttt{FRRN}-based MoE~\cite{pavlitskaya2020using}. RU stands for a residual unit.}
    \end{subfigure}
    \caption{MoE architectures of the \texttt{DeepLabv3$+$} and \texttt{FRRN}-based models.}
    \label{fig:architectures}
\end{figure*}

\section{Approach}

To investigate the adversarial robustness of MoE models, we use the architecture from our previous works~\cite{pavlitskaya2020using,pavlitskaya2022evaluating} as a representative example and apply per-instance and universal attacks in white-box and transfer settings.

\subsection{MoE architecture}
The MoE architecture contains two experts, pre-trained on two disjoint data subsets, a trainable gate, and optionally an additional convolutional layer (see Fig. \ref{fig:architectures}). The gate is a small neural network containing one convolutional and two fully connected layers. It gets the concatenation of the features from experts as input and uses \textit{softmax} to predict expert weights. We consider two gate architectures: a \textit{simple} gate predicts one weight per expert, whereas a \textit{classwise} gate predicts a weight per class per expert. 

Furthermore, we consider adding an additional convolutional layer after the weighted sum of expert predictions and before the final softmax (see Figure~\ref{fig:architectures}). In total, there are, therefore, four MoE architectures.

\subsection{Threat Model}

We evaluate \textit{fast gradient sign method} (FGSM)~\cite{goodfellow2014explaining} and \textit{projected gradient descent} (PGD) attacks~\cite{madry2018towards}. Both methods are white-box attacks, i.e., an attacker has full access to model architecture, weights, and training data. Furthermore, both attacks are performed in an untargeted manner, i.e., we do not target a specific class but instead maximize an error on a ground-truth class. 

FGSM is a one-step attack, where for an input $x$ and the ground truth label $y$, the adversarially perturbed input $x_{adv}$ is obtained as follows:

\begin{equation}
    x^{adv} = x + \delta = x + \epsilon \cdot sign (\nabla_x L (\Theta,x,y),
\end{equation}
where $\delta$ is the adversarial perturbation, $L$ is the loss function used for training of a neural network with parameters $\Theta$, and $\epsilon$ is the magnitude of perturbation, s.t.  $||\delta||_\infty \leq \epsilon$, i.e. the norm of the perturbation is bound to be at most $\epsilon$. 

In an iterative version of FGSM, also called \textit{basic iterative method} (BIM)~\cite{kurakin2017adversarial}, the perturbed input for iteration $t+1$ is computed as follows:
\begin{equation}
\begin{split}
    x^{adv}_{t+1} = clip_{x,\epsilon}(x^{adv}_{t} + \delta_{t+1})  \\
    = clip_{x,\epsilon}(x^{adv}_{t} + \alpha \cdot sign (\nabla_x L (\Theta,x^{adv}_{t},y)),
\end{split}
\end{equation}
where $\alpha$ is a learning rate and the $clip$ function ensures that the perturbed image remains within the $L_\infty$ ball around $x$. 

The PGD is similar to BIM, but starts at a uniformly random point within the $L_\infty$ ball around a sample $x$, whereas BIM initializes at the original point. As in several previous works~\cite{uesato2018adversarial,carlini2017towards}, we replace the original vanilla gradient update in PGD with Adam~\cite{kingma2014adam} to ensure faster convergence.

In addition to per-instance attacks, where the perturbation is computed for each input image, we also perform PGD attacks in a universal manner~\cite{moosavi2017universal}, s.t. a single attack pattern is generated for the whole dataset.

\section{Experiments and Evaluation}
\label{sec:experiments}

\subsection{Experimental Setup}

\textbf{Models:} We evaluate the approach using two different expert base architectures: the Full-Resolution Residual Network (\texttt{FRRN})~\cite{pohlen2017full} and \texttt{DeepLabv3$+$}~\cite{chen2018encoder} with the ResNet-101 backbone~\cite{he2016deep}, both pre-trained on ImageNet.  For the \texttt{FRRN}-based models, we use FRRN-A, a shallower version of FRRN. The MoE uses $\text{FRRU}_{384}$ as a feature extraction layer (see Figure~\ref{fig:architectures}). For the \texttt{DeepLabv3$+$}-based models, we extract features from the atrous spatial pyramid pooling (ASPP) layer, the last decoder layer.

\textbf{Dataset:} For the expert training, we use the manual A2D2~\cite{geyer2020a2d2} data split as defined in \cite{pavlitskaya2020using}. \textit{Highway} and \textit{urban} subsets contain 6,132 training, 876 validation, and 1,421 test samples each. We report results on the combined \textit{highway-urban} dataset comprising 2,842 images. Images in the A2D2 dataset have the original resolution $1280\times1920$ pixels, and they are processed at the resolution $912\times912$ for the \texttt{DeepLabv3$+$} and $480\times640$ for the \texttt{FRRN} models. We use the complete label set of the A2D2 dataset here, comprising 38 classes, analogously to~\cite{pavlitskaya2022evaluating}. Each expert is trained on the corresponding subset of data; the baseline has the same architecture as the experts, and is trained on the combined \textit{highway-urban} training data. The MoE is also trained on the combined \textit{highway-urban} data.

\begin{figure}[t]
\centering
\begin{subfigure}[t]{\linewidth}
    \includegraphics[width=\linewidth]{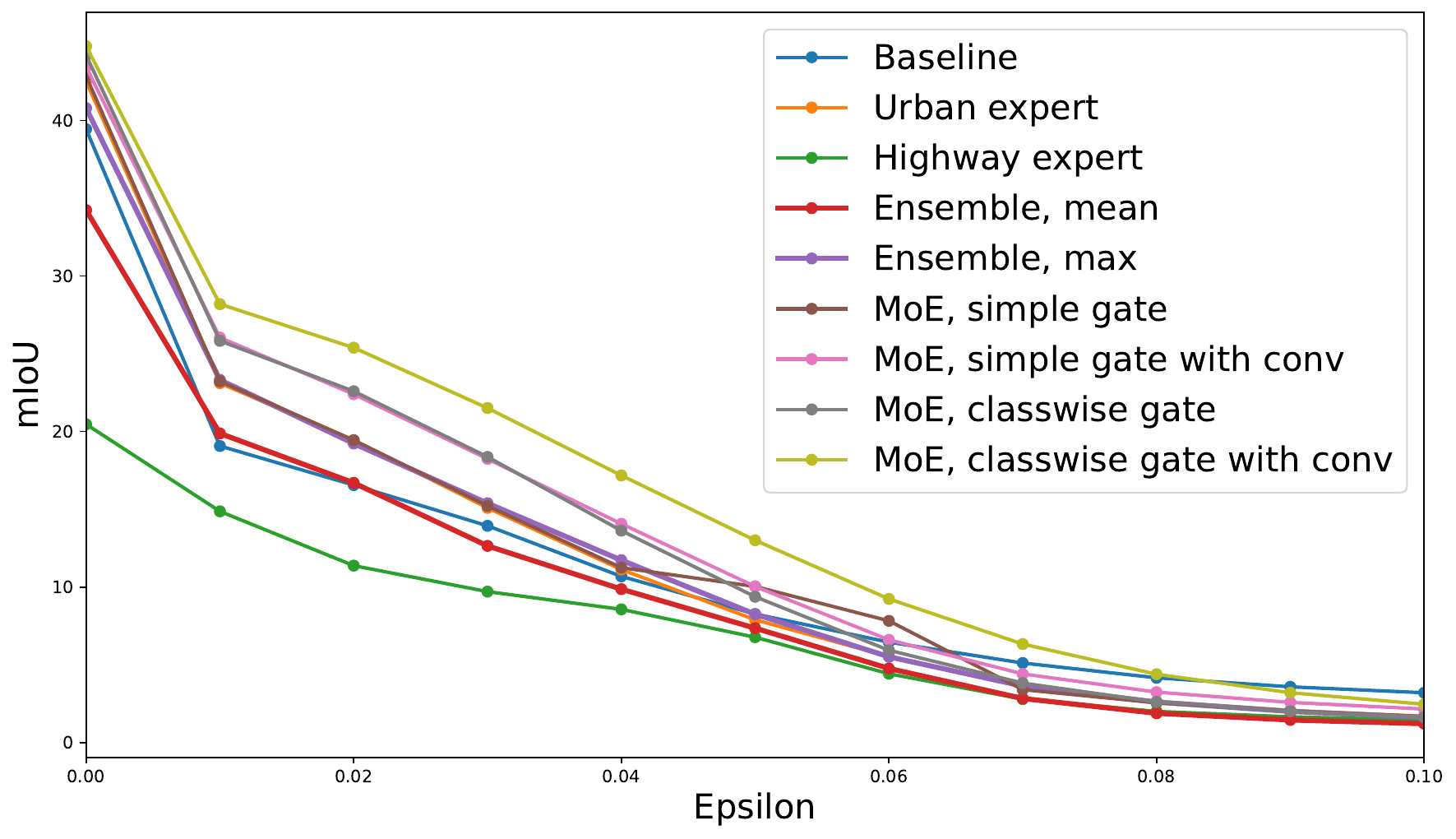}
  	\caption{\texttt{DeepLabv3$+$}-based models}
\end{subfigure}
\begin{subfigure}[t]{\linewidth}
 \includegraphics[width=\textwidth]{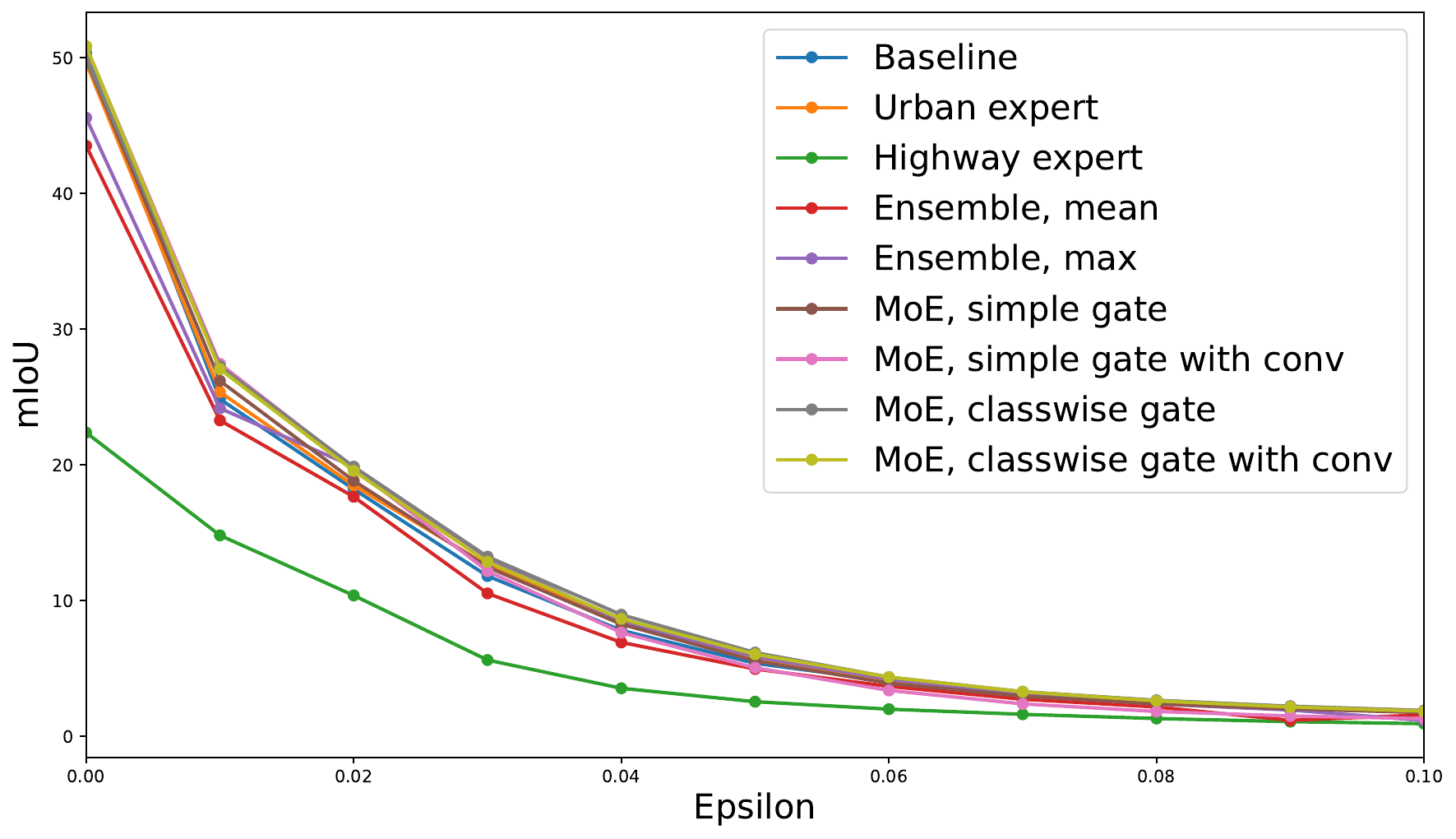}
 	\caption{\texttt{FRRN}-based models}
\end{subfigure}
    \caption{Performance of \texttt{DeepLabv3$+$}-based experts, ensembles, and MoEs under FGSM attacks of different strength.}
    \label{fig:fgsm}
\end{figure}

\textbf{Training Setup:} The experts and the baseline are trained for 200 epochs with a batch size of two, and the MoE is trained for 100 epochs with a batch size of six. The training uses SGD with the polynomial learning rate decay with an initial value 0.01. All trainings were performed on an NVIDIA RTX 2080 Ti GPU with 11GB VRAM.

\textbf{Metrics:} We use the standard segmentation evaluation metric mIoU to evaluate the performance of segmentation models. Higher mIoU under attack would indicate better robustness to adversarial attacks, and low mIoU for transfer attacks would mean high transferability. All attacks are evaluated on the corresponding test datasets. For universal attacks, the noise pattern is trained using the train and validation subsets and evaluated on the respective test data. 

\textbf{Attack Settings:} We use FGSM with different $\epsilon$ values as well as PGD-10 with a learning rate 0.01 and Adam optimizer~\cite{kingma2014adam}.

\begin{table*}[t]
\caption{Experts and the MoE under attack, $\epsilon=0.05$. The best-performing and the most robust models among the single-model and combination approaches are marked.}
\centering
  %\resizebox{1.0\linewidth}{!}{
    \begin{tabular}{|r| c | c | c | c  |  }
    \hline
    \textbf{Model} & \textbf{No attack} & \textbf{FGSM}  & \textbf{PGD-10}  &  \textbf{Universal PGD-10}\\ \hline
    \rowcolor{LightBlue}
    \multicolumn{5}{|c|}{\textbf{\texttt{FRRN}-based models}}\\ 
    Baseline & \textbf{50.32} & 5.38 (-89.31\%) & 0.35 (-99.30\%)  &\textbf{6.04} (\textbf{-87.99\%})  \\ 
    Urban expert &   49.63   & \textbf{5.80} (\textbf{-88.31\%}) &  \textbf{0.55} (-98.89\%)& 4.53 (-90.87\%)\\
    Highway expert &   22.38   & 2.55 (-88.61\%) & 0.38 (\textbf{-98.30\%}) &  2.67 (-88.09\%)\\ \hline
    Ensemble (mean)  & 43.52 & 4.96 (-88.60\%)  & 0.55 (-98.73\%)& 4.01 (-90.79\%)\\
    Ensemble (max) & 45.58 & 5.86 (\textbf{-87.14\%}) & 0.45 (-99.02\%)&  \textbf{6.08} (\textbf{-86.65\%})\\
    MoE, simple gate& 49.91 & 5.56 (-88.86\%)& 0.57 (-98.86\%) & 4.16 (-91.66\%)\\
    MoE,  simple gate and conv  &  50.81 & 5.05 (-90.06\%) & 0.40 (-99.21\%) &   3.39 (-93.33\%)\\ 
    MoE, classwise & 50.00 & \textbf{6.17} (-87.66\%) & \textbf{0.75} (\textbf{-98.50\%})& 4.46 (-91.08\%)\\
    MoE, classwise gate and conv &  \textbf{50.85}  & 6.05 (-88.10\%)& 0.66 (-98.71\%)& 3.82 (-92.49\%)\\\hline

    %%% DeepLab Models
    \rowcolor{LightBlue}
    \multicolumn{5}{|c|}{\textbf{\texttt{DeepLabv3$+$}-based models}}\\ 
    Baseline & 39.44 & \textbf{8.27} (-79.03\%) & 1.04 (-97.36\%) &  3.60 (-90.87\%) \\ 
    Urban expert &  \textbf{42.51} & 7.90 (-81.42\%) & \textbf{1.10} (-97.41\%) & \textbf{4.70} (-88.94\%) \\ 
    Highway expert &  20.47 & 6.78 (\textbf{-66.88\%})  & 0.80 (\textbf{-96.09\%})& 3.70 (\textbf{-81.92\%})\\ 
    \hline
    Ensemble (mean) & 34.23 & 7.36 (-78.50\%)  & 1.28 (-96.26\%) & 1.20 (-96.49\%) \\
    Ensemble (max) & 40.79 & 8.26 (-79.75\%) & 0.83 (-97.97\%) & 1.70 (-95.83\%) \\
    MoE, simple gate & 42.85 & 7.83 (-81.73\%) & 1.20 (-97.20\%) & 3.30 (-92.30\%) \\
    MoE, simple gate and conv &  43.43 & 10.04  (-76.88\%) & 1.73 (-96.02\%) & \textbf{3.40} (\textbf{-92.17\%})\\ 
    MoE, classwise gate &  \textbf{44.35} & 9.38 (-78.85\%)  & 1.32 (-97.02\%) & 1.20 (-97.29\%) \\
    MoE, classwise gate and conv & 44.26  & \textbf{13.01} (\textbf{-70.61\%})  & \textbf{2.78} (\textbf{-93.72\%}) & 1.10  (-97.51\%) \\
    \hline
    \end{tabular}
%}
\label{tab:attacks}
\end{table*}

\begin{figure*}[t]
\centering
\begin{subfigure}[t]{\columnwidth}
  \includegraphics[width=1.0\textwidth]{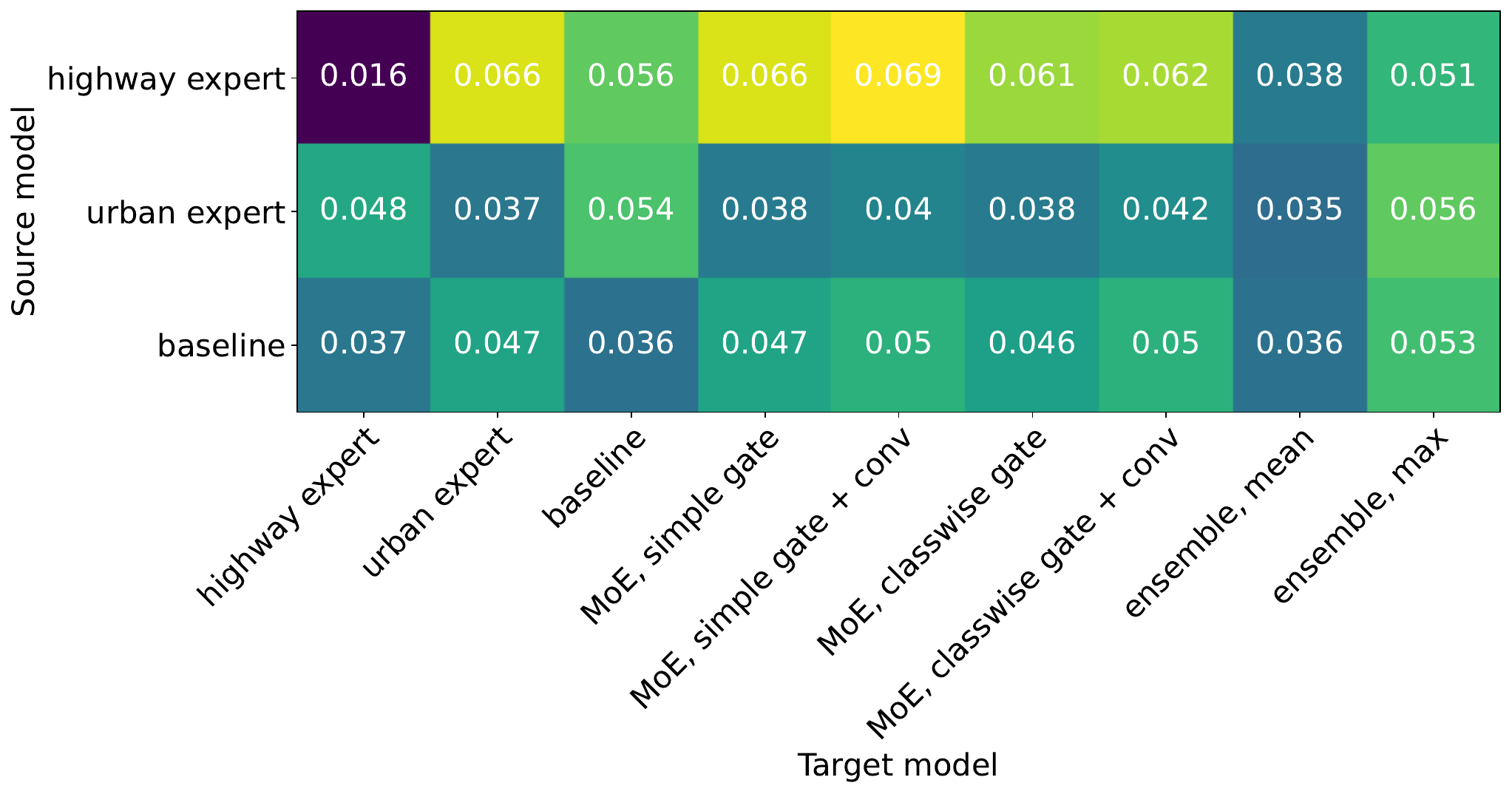}
  	\caption{\texttt{DeepLabv3$+$}-based models}
\end{subfigure}
\begin{subfigure}[t]{\columnwidth}                              
  \includegraphics[width=1.0\textwidth]{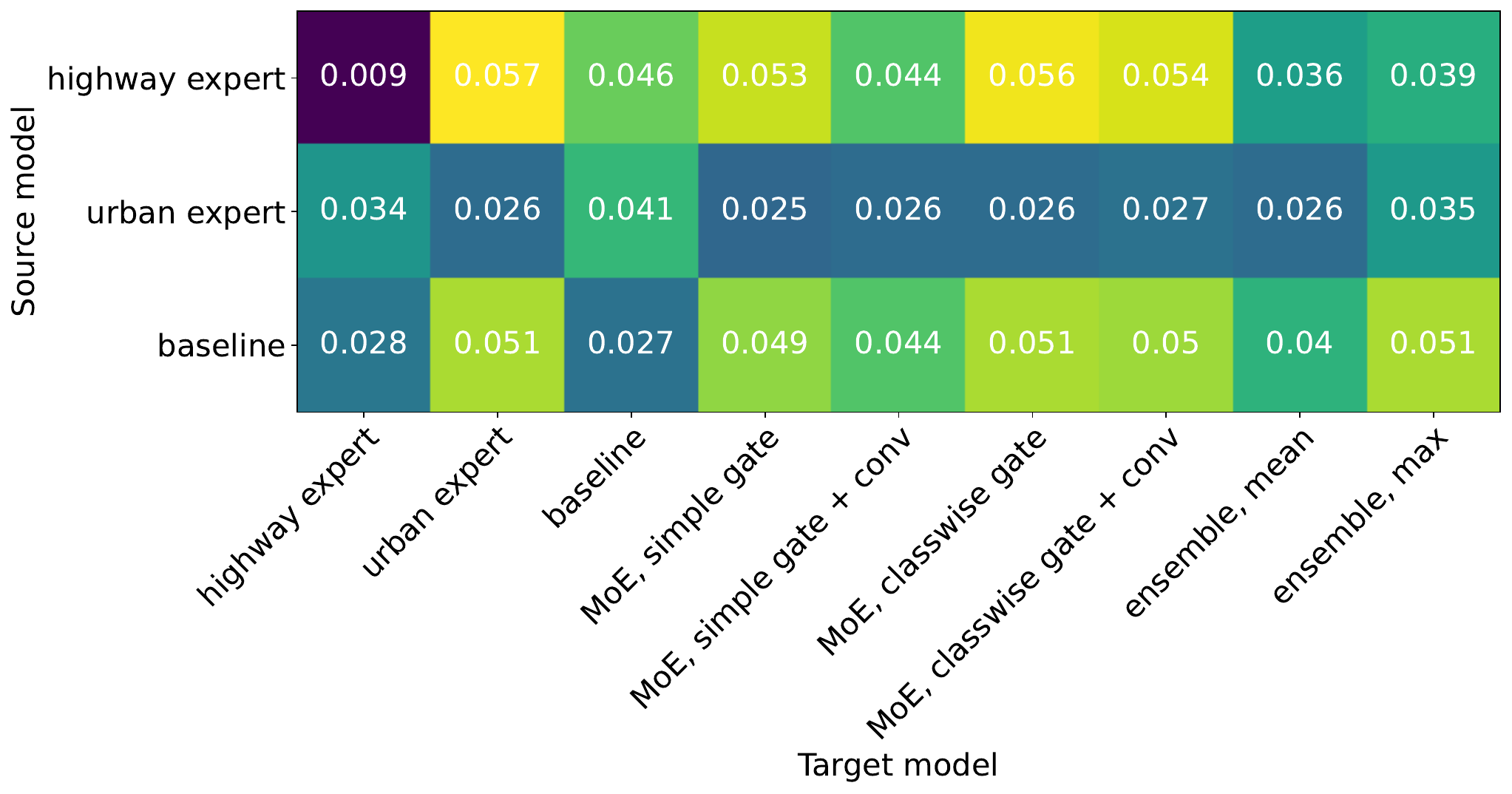}
  	\caption{\texttt{FRRN}-based models}
\end{subfigure}
    \caption{Transfer PGD attacks, $\epsilon=0.05$. }
    \label{fig:transfer_attack}
\end{figure*}

\subsection{Performance under White-box Attacks}

We evaluate the robustness of individual experts, the baseline, and four MoE architectures, originally proposed in~\cite{pavlitskaya2020using}. In addition to MoEs, we also evaluated the ensembles of two experts. We combined the classwise probabilities output of experts by taking either the mean or the maximum of the expert predictions. As Figure~\ref{fig:fgsm} shows, in FGSM experiments, the MoEs with an additional convolutional layer tend to retain higher mIoU longer, especially for larger $\epsilon$ values, whereas MoE architectures without an additional convolutional layer show a faster drop in mIoU. Overall, the MoEs with a classwise gate have demonstrated a smaller drop in accuracy under all attacks (see Table~\ref{tab:attacks}). For the \texttt{DeepLabv3$+$}-based models, the MoE with a classwise gate and an additional convolutional layer showed the best robustness. Furthermore, \texttt{DeepLabv3$+$} models were more robust to attacks than \texttt{FRRN} models. Note that due to the re-implementation of model architectures, the performance of models without attacks slightly differs from~\cite{pavlitskaya2022evaluating}.

Furthermore, we have also compared the performance of models under attack on \textit{highway} and \textit{urban} data. We have observed that the highway expert has the worst robustness on data beyond its domain but can retain robustness on the highway data. The urban expert demonstrates behavior similar to the baseline expert on the urban and combined dataset, but it is less robust than the baseline on the highway data.

\subsection{Performance under Transfer Attacks}
Universal PGD-generated adversarial noise has demonstrated good transferability between models (see Figure~\ref{fig:transfer_attack}). Adversarial noise generated for a highway expert has led to the weakest attacks, whereas that for the urban expert has led to attacks stronger than the baseline. MoEs with an additional convolutional layer have demonstrated better transfer attack robustness than those without an extra layer. 

\section{Conclusion}
In this work, we have explored the adversarial robustness of the mixture-of-experts architectures at the model level on the semantic segmentation task. While model combination approaches like ensembles have been studied in detail previously, mixtures of experts have received significantly less attention so far. We have compared experts, ensembles, and four types of MoEs based on \texttt{FRRN$+$} and \texttt{DeepLabv3$+$} architectures under per-instance and universal FGSM and PGD attacks. Our experiments have shown that, in most cases, the MoE-based models exhibit a lower drop in accuracy compared to ensembles. Especially an MoE with an additional convolutional layer has demonstrated the smallest drop in accuracy, especially for the \texttt{DeepLabv3$+$}-based architectures. Furthermore, our experiments with transfer attacks have shown that MoEs can better withstand attacks of this type.

\section*{Acknowledgment}

This work was supported by funding from the Topic Engineering Secure Systems of the Helmholtz Association (HGF) and by KASTEL Security Research Labs (46.23.03).

{\small
\bibliographystyle{IEEEtran}
\bibliography{references}
}
\end{document}